\documentclass[10pt,twocolumn,letterpaper]{article}

\usepackage{cvpr}      
\usepackage{colortbl}
\usepackage{times}
\usepackage{makecell}
\usepackage{epsfig}
\usepackage{amsmath}
\usepackage{amssymb}
\usepackage{multirow}
\usepackage{mathrsfs}
\usepackage{tabulary}
\usepackage[dvipsnames]{xcolor}
\usepackage{colortbl}
\definecolor{mygray}{gray}{0.6}
\definecolor{mygray-bg}{gray}{0.9}
\usepackage{graphicx}
\usepackage{amsmath}
\usepackage{amssymb}
\usepackage{booktabs}

\usepackage[pagebackref,breaklinks,colorlinks]{hyperref}

\newcommand{\tablestyle}[2]{\setlength{\tabcolsep}{#1}\renewcommand{\arraystretch}{#2}\centering\footnotesize}

\newcommand{\app}{\raise.17ex\hbox{$\scriptstyle\sim$}}

\newcolumntype{x}[1]{>{\centering\arraybackslash}p{#1pt}}
\newlength\savewidth\newcommand\shline{\noalign{\global\savewidth\arrayrulewidth\global\arrayrulewidth 1pt}\hline\noalign{\global\arrayrulewidth\savewidth}}

\def\cvprPaperID{1524} 
\def\confName{CVPR}
\def\confYear{2022}

\begin{document}

\title{Reinforced Structured State-Evolution for Vision-Language Navigation}
\author{Jinyu Chen$^{1,2}$, Chen Gao$^{1,2}$, Erli Meng$^{3}$, Qiong Zhang$^{3}$, Si Liu$^{1, 2}$\thanks{Corresponding author: \textit{Si Liu}.}~\\
\small{$^1$Institute of Artificial Intelligence, Beihang University} \\ 
\small{$^2$Hangzhou Innovation Institute, Beihang University} \\ 
\small{$^3$Xiaomi AI Lab, Xiaomi Inc}\\
\small\url{https://github.com/chenjinyubuaa/SEvol}
}

\maketitle

\begin{abstract}
Vision-and-language Navigation (VLN) task requires an embodied agent to navigate to a remote location following a natural language instruction. 
Previous methods usually adopt a sequence model (e.g., Transformer and LSTM) as the navigator. In such a paradigm, the sequence model predicts action at each step through a maintained navigation state, which is generally represented as a one-dimensional vector. 
However, the crucial navigation clues (i.e., object-level environment layout) for embodied navigation task is discarded since the maintained vector is essentially unstructured. 
In this paper, we propose a novel Structured state-Evolution (SEvol) model to effectively maintain the environment layout clues for VLN. Specifically, we utilise the graph-based feature to represent the navigation state instead of the vector-based state. Accordingly, we devise a Reinforced Layout clues Miner (RLM) to mine and detect the most crucial layout graph for long-term navigation via a customised reinforcement learning strategy. Moreover, the Structured Evolving Module (SEM) is proposed to maintain the structured graph-based state during navigation, where the state is gradually evolved to learn the object-level spatial-temporal relationship. 
The experiments on the R2R and R4R datasets show that the proposed SEvol model improves VLN models' performance by large margins, e.g.,  $+3\%$ absolute SPL accuracy for NvEM and $+8\%$ for EnvDrop on the R2R test set.
\end{abstract}
\section{Introduction}                                        
\label{sec:intro}
\begin{figure}
\vspace{5mm}
   \begin{center}
      \includegraphics[width=1\linewidth]{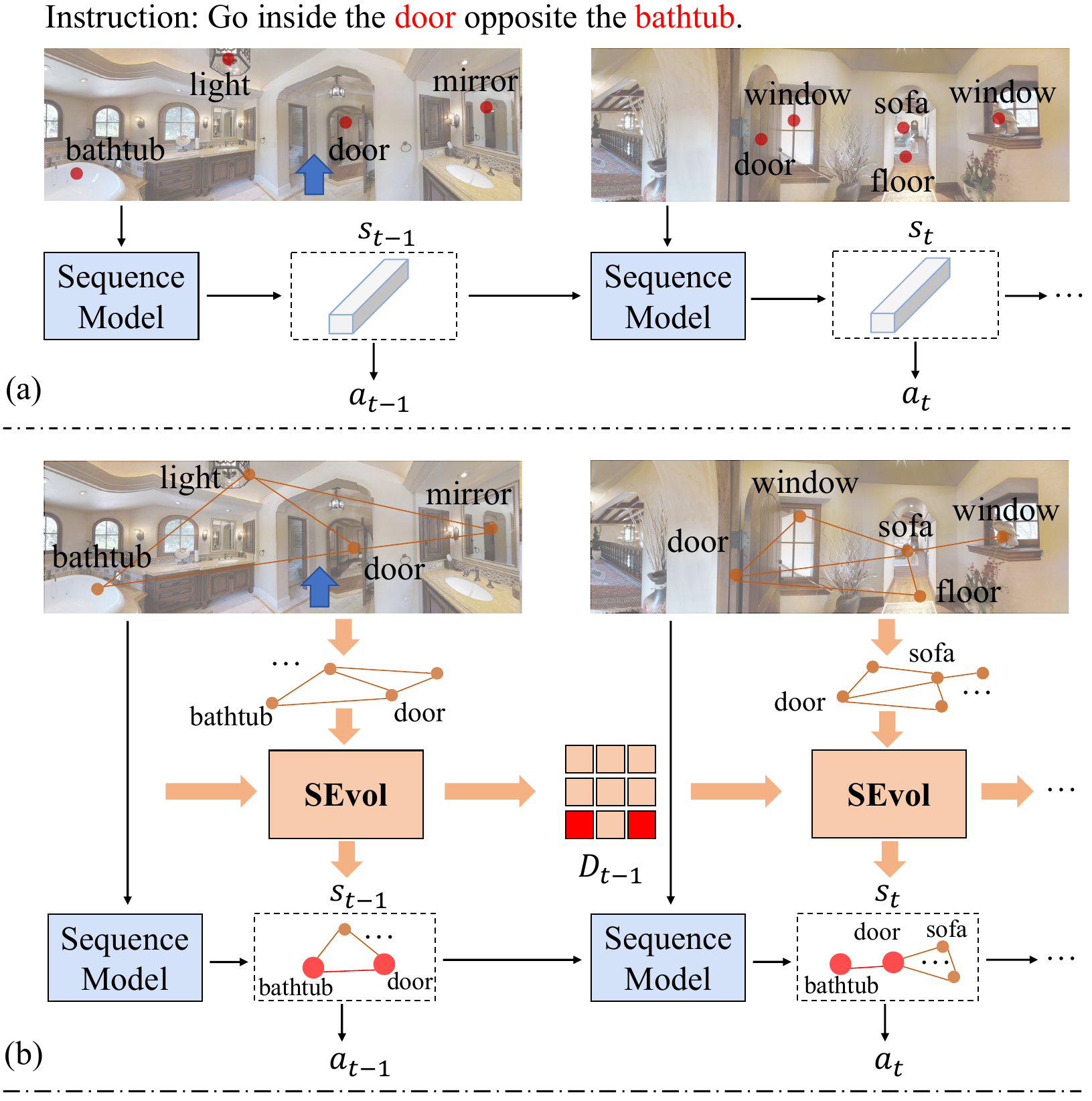}
   \end{center}\vspace{-4mm}
      \caption{At each step $t$, (a) previous methods predict action $a_t$ based on a \emph{vector-based} navigation state $s_t$ while the object-level layout memory is discarded; (b) we propose SEvol to maintain a \emph{graph-based} navigation state $s_t$, which can effectively record the layout memory via the structured state-evolution.
      }
   \label{fig:fig1}
\end{figure}

In recent years, Embodied-AI (E-AI) that requires embodied agents to complete tasks has arrested extensive interests of both computer version and natural language processing community. Numerous datasets ~\cite{chang2017matterport3d,Savva_2019_ICCV} have been constructed to simulate realistic environments to support various embodied tasks such as navigation ~\cite{Savva_2019_ICCV,xiazamirhe2018gibsonenv}, interactive learning~\cite{robothor,ALFRED20} and multi-agent cooperation~\cite{mordatch2018emergence}, \etc. 

One of the most attractive application scenarios of E-AI is the Vision-and-Language Navigation (VLN) task~\cite{anderson2018vision}, where the goal is for an embodied agent in a 3D environment to navigate to the specific location following the natural language instruction. As shown in Figure~\ref{fig:fig1}(a), previous methods~\cite{ma2019regretful,tan2019learning,ma2019self,fried2018speaker} usually adopt sequence model (\eg, Transformer and LSTM) to model the sequential decision process. At each step $t$, the action is predicted according to the navigation state $s_t$, which maintains the historical and current environment information. 
Commonly, the navigation state is maintained in the form of \emph{unstructured one-dimensional vector}. The environment clues at each step, \ie, visual and orientation features, are all compressed and pooled into this unstructured vector. Therefore, the \emph{structured object-level environment layout} information is discarded during this process. However, the environment layout clues are crucial for the embodied VLN task. As shown in Figure~\ref{fig:fig1}(a), to fulfill the instruction like `\textit{go inside the door opposite the bathtub}', the agent of previous methods is confused at step $t$ since the landmark object `\textit{bathtub}' can not be observed, and the agent needs to utilise the historical layout clues (\ie, the door is opposite the bathtub) at step $t-1$ to make the right action.

Therefore, we aim to improve the VLN paradigm via maintaining a structured navigation state, in which three important factors need to be considered. (\romannumeral1) How to represent the navigation state to contain structured layout memory. (\romannumeral2) How to mine the pivotal layout information for the current and future decisions according to the instruction. (\romannumeral3) How to store and update a structured state while satisfying the property of long short-term memory.

To achieve the aforementioned objectives, we propose a Structured state-Evolution (SEvol) model as shown in Figure~\ref{fig:fig1}(b), where multi-fold innovations are made.
(\romannumeral1) Instead of the \emph{vector-based} feature, we propose to adopt the \emph{graph-based} feature as the navigation state, which is capable of holding a structured layout memory. 
(\romannumeral2) We design a Reinforced Layout clues Miner (RLM) to mine the most crucial layout information. RLM learns to detect and sample the essential subgraph from the whole layout graph, conditioned on both the current navigation state and the instruction, \eg, sampling subgraph $<$\emph{door}, \emph{opposite}, \emph{bathtub}$>$ according to the language `\textit{go inside the door opposite the bathtub}'. In RLM, we customise a reinforcement learning strategy to make the miner focus on both the immediate interests and the long-term influence of the subgraph sampling.
(\romannumeral3) To effectively store and update a structured graph-based state during the whole navigation process, we devise a Structured Evolving Module (SEM). Specifically, SEM takes the current graph features from RLM as input to evolve the navigation state at each step. The evolution of the navigation state is achieved by interacting with a learnable matrix $D$ (shown in Figure~\ref{fig:fig1}(b)), which stores the structured layout memory. $D$ is updated through a matrix version of recurrent neural network.
Thus the navigation state contains object-level spatial-temporal relationship that assists the action decision, \eg, the relation  `\emph{door opposite bathtub}' in Figure~\ref{fig:fig1}(b).
Experiments on Room-to-Room (R2R)~\cite{anderson2018vision} and Room-for-Room (R4R) show that the proposed SEvol model improves VLN models' performance by large margins.

To summarise, we make the following contributions:
\vspace{-4pt}
\begin{itemize}
\setlength\itemsep{-1pt}
    \item We propose a simple yet effective SEvol model that provides new insights to the VLN community. The structured navigation state is leveraged to maintain the object-level environment layout during navigation. SEvol achieves state-of-the-art performance on R2R.
    \item We design a Reinforced Layout clues Miner (RLM) to learn how to detect and sample the most critical subgraph features from the layout graph for current and future action decisions. 
    \item We devise a Structured Evolving Module (SEM) to gradually evolve the structured navigation-state along with the navigation process, maintaining a long short-term layout memory.
\end{itemize}

\section{Related Work}
\noindent\textbf{Vision-and-Language Navigation~(VLN).}
Learning to conduct navigation in a simulated photo-realistic environment following the human-annotated nature language instruction, \ie, VLN task~\cite{anderson2018vision,sotp2019acl}, has drawn significant interests from both academic and industry fields in recent years. Numerous methods~\cite{fried2018speaker,ma2019self,ma2019regretful,tan2019learning,hu2019you,wang2020active,hao2020prevelant,zhu2020vision,hong2020recurrent,wang2021structured,Gubur2021AirBert,qi2021road,liu2021vision} have been proposed for the VLN task. Early work Speaker-Follower~\cite{fried2018speaker} designs an instructions argumentation strategy, and EnvDrop~\cite{tan2019learning} increases the visual diversity of the synthesised training samples. Besides, ~\cite{ma2019self,zhu2020vision} introduce auxiliary losses to further improve the ability of cross-modal understanding, and~\cite{wang2021structured,deng2020evolving} employ trajectory-graph to keep the global navigation memory.

Recently, most VLN works focus on how to utilise a more powerful transformer-based vision-language model to improve performance. CKR \cite{Gao_2021_CVPR} adopts the transformer decoder to model the sequential navigation process. VLN-BERT~\cite{majumdar2020improving} and Airbert~\cite{Guhur_2021_ICCV} leverage the vision-language transformer pre-trained on other large-scale vision-language datasets~\cite{sharma2018conceptual} to conduct instruction-trajectory matching. Other works~\cite{hong2020recurrent,qi2021road} focus on customising a transformer-based model, which is specifically tailored for the VLN task and can inherit the capability from pre-training models. However, the previous VLN methods pay less attention to the issue of \emph{unstructured navigation state}, which unintentionally reduces the crucial navigation clues, \ie, object-level environment layout.
\begin{figure*}[t]
   \begin{center}
      \includegraphics[width=0.93\linewidth]{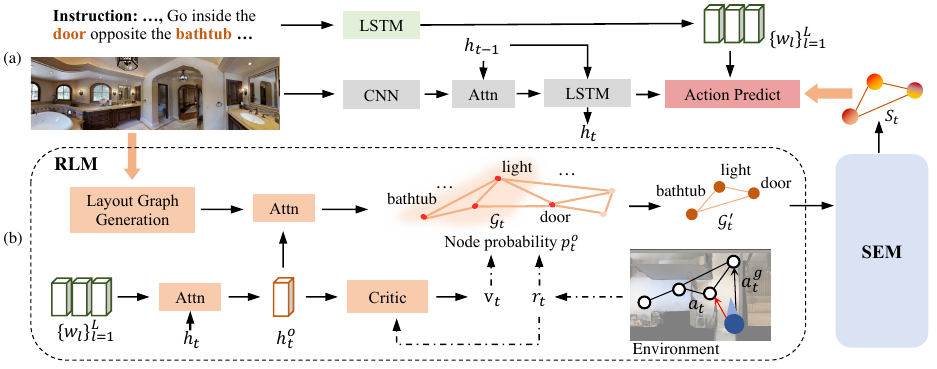}
   \end{center}
   \vspace{-5mm}
   \caption{\textbf{The Overall Pipeline.} (a) The typical paradigm of VLN methods. (b) The proposed SEvol contains two components, \ie, RLM and SEM. RLM digs out the pivotal layout graph according to the instruction via a customised reinforcement learning strategy. SEM evolves the structured navigation state to maintain the crucial layout memory during navigation.}
\label{fig:pipline}
\vspace{-4mm}
\end{figure*}

\noindent\textbf{Dynamic Graph Neural Networks.}
In the field of data mining, the dynamic graph neural network is an effective tool to mine the information from a sequence of graph-based data that changes over time (\eg, social networks). Some approaches~\cite{perozzi2014deepwalk,yu2018netwalk} are designed to extract the spatial-temporal relationship from the dynamic graph. \cite{trivedi2017know,goyal2020dyngraph2vec} take the dynamic graph as the continuous changes over the initial graph, which are advantageous for the event time prediction. Besides,~\cite{deng2020evolving,MDHN2021} take several frames of the graph and leverage GCN~\cite{kipf2016semi}/sequential model~\cite{chung2014empirical} to extract features from the graph that changes fast over time.
Inspired by the area of data mining, we propose SEM to maintain and update a graph-based navigation state, where different object-level layouts can be extracted dynamically.
\section{Method}
\subsection{Problem Setup and Overview}
\noindent\textbf{Problem Setup.}
\label{sec:setup}
In VLN~\cite{anderson2018vision}, the agent is required to reach the described locations following the instruction. At each step $t$, the agent observes a panoramic RGB view, which is further divided into 36 discrete views $\{v_{t,i}\}_{i=1}^{36}$. Each view is represented via a RGB image $v_{t,i}$ with its orientation information (heading $\theta_{t,i}$, elevation $\psi_{t,i}$), where $i$ is the view index. Generally, each view's feature $f_{t,i}$ is obtained through $f_{t,i} = [\mathcal{F}(v_{t,i}), \mathcal{E}(\theta_{t,i},\psi_{t,i})]^\top$, where $[\cdot,\cdot]$ denotes concatenation,  $\mathcal{F}(\cdot)$ is the image feature extractor and $\mathcal{E}(\cdot)$ represents the orientation embedding function~\cite{tan2019learning}, which is defined as $[cos(\theta_{t,i}), sin(\theta_{t,i}), cos(\psi_{t,i}), sin(\psi_{t,i})]$.
Besides, there are $K_t$ candidate views at each step $t$ that are navigable. The agent needs to take an action $a_t$, \ie, selecting one from $K_t$ candidates to move to that location.

\noindent\textbf{Overview.}
The basic pipeline of VLN methods~\cite{an2021neighborview,tan2019learning} is shown in Figure~\ref{fig:pipline}(a). The proposed SEvol (shown in Figure~\ref{fig:pipline}(b)) is set as an additional branch for providing a structured navigation state. SEvol consists of two components, \ie, Reinforced Layout clues Miner (RLM) and Structured Evolving Module (SEM).
The basic VLN pipeline (Figure~\ref{fig:pipline}(a)) leverages a bi-directional LSTM~\cite{huang2015bidirectional} to encode the instruction, and obtains the word-level language feature $\{w_l\}^L_{l=1}$, where $L$ is length. $w_L$ is the sentence level feature and serves as the initial hidden state of the LSTM decoder. At each navigation step $t$, current visual feature $\{f_{t,i}\}^{36}_{i=1}$ is fed to the LSTM decoder to update its hidden state $h_t \in \mathbb{R}^{N_h \times 1}$:
\begin{equation}
   \left\{\begin{matrix}
   \begin{aligned}
    \tilde{f}_t &= {attn}(h_{t-1},\{f_{t,i}\}^{36}_{i=1}); \\
    h_t &=  {LSTM}([\tilde{f}_t,a_{t-1}],h_{t-1}).
    \label{eq:LSTM-update}
    \end{aligned}
   \end{matrix}\right.
\end{equation}
Note that we represent the normal attention mechanism as $attn(\cdot,\cdot)$ in the paper. For example, $attn(x,Y) = softmax( x^\top W_a Y^\top )Y $, where $x\in \mathbb{R}^{N_1 \times 1}$, $Y \in \mathbb{R}^{N_2 \times {N_3}}$, and $W_a \in \mathbb{R}^{N_1 \times N_3}$ are trainable parameters.

At each step $t$, SEvol firstly produces layout graph $\mathcal{G}_t$ from the visual observation.  As shown in Figure~\ref{fig:pipline}(a), then RLM aims to utilise the language feature $\{w_l\}^L_{l=1}$ (\eg, `\textit{go inside the door opposite the bathtub}') to mine the key subgraph $\mathcal{G}_t'$ (\eg, `\textit{bathtub}' and `\textit{door}') from $\mathcal{G}_t$. The whole process is optimised through a customised reinforced learning strategy since the RL-based objective can make the miner consider both current and future influences. 
The reward $r_t$ for the RLM is based on whether the agent moves close or reaches the target position.
Next, SEM takes current $\mathcal{G}_t'$ from RLM and layout memory $D_{t-1}$ as inputs to evolve the structured navigation state $S_t$, where $D_{t-1}$ is an iterable matrix that is used to record the historical layout memory. The evolved $S_t$ is further adopted to predict the final action $a_t$. In the following, we detailly introduce RLM in Section~\ref{sec:REM} and SEM in Section~\ref{seg:SEM}, respectively.

\subsection{Reinforced Layout clues Miner}
\label{sec:REM}
\noindent\textbf{Layout Graph Generation.}
\label{sec:gen}
To extract the object-level environment layout, at each step $t$, we detect top $K$ salient objects $ \mathcal{O}_t=\{\textbf{o}_{t,k}\}^{K}_{k=1}$ via the Faster R-CNN~\cite{ren2015faster}, where $\textbf{o}_{t,k}$ represents the object entity. We generate a fully connected layout graph $\mathcal{G}_t=\{\mathcal{O}_t,\mathcal{A}_t\}$ base on the object set, where $\mathcal{A}_t$ is the edge set of $\mathcal{G}_t$. 
To encode the object node set $\mathcal{O}_t$ into a node feature matrix $O_t$, we consider the semantic and relation position information. The k-th object feature $O_{t,k}$ is defined as:
\begin{equation}
O_{t,k} = [G(\textbf{o}_{t,k}), \mathcal{E}(\theta_{\textbf{\textbf{o}}_{t,k}},\psi_{\textbf{o}_{t,k}})]^\top,
   \label{eq:object-features}
\end{equation}

where $G(\cdot)$ donates the GloVe~\cite{pennington2014glove} embedding of the object's label. $\theta_{\textbf{o}_{t,i}}$ and $\psi_{\textbf{o}_{t,i}}$ are the heading and elevation of the object relative to the current direction of agent. $\mathcal{E}(\cdot)$ is the orientation embedding function defined in Section~\ref{sec:setup}. 
Then based on the spatial relationships between objects, we define the adjacency matrix $A_t$ as:
\begin{equation}
A_{t,[i,j]} = d(\theta_{\textbf{o}_{t,i}} - \theta_{\textbf{o}_{t,j}}),\\
   \label{eq:adge-generate}
\end{equation}
where $d(\cdot)$ is a decrease function of the heading difference and is defined in Section~\ref{sec:exp}.

\vspace{0.5mm}
\noindent\textbf{Subgraph Mining.} The generated layout graph $\mathcal{G}_t$ contains noise (\eg, navigation-irrelevant object relations). Therefore, to focus on the most pivotal layout information, we propose the RLM to sample a representative subgraph containing $M$ object nodes, as shown in Figure~\ref{fig:pipline}(b). Intuitively, the mining process should be dependent on the instruction. Thus we leverage the language-aware feature $\tilde{h}_t=attn(h_t,\{w_l\}^{L}_{l=1})$ to compute the object importance for the subgraph mining:
\begin{equation}
    p^o_t =  {softmax}({h_t^o}^\top  W_2  {O_t}^\top),\quad h_t^o = \delta( W_1 \tilde{h}_t ),
\end{equation}
where $W_1\in \mathbb{R}^{N \times N_h}$ and $W_2\in \mathbb{R}^{N \times N_o}$ are learnable parameters. $N=100$ denotes the hidden size and $\delta(\cdot)$ is the ReLU activation function. $p_t^o \in \mathbb{R}^{1 \times K}$ indicates the object's importance during navigation process. Sampling the subgraph $\mathcal{G}_t'=\{\mathcal{O}_t', \mathcal{A}_t'\}$ from $\mathcal{G}_t$ includes two steps. \emph{Firstly}, we select the top $M$ object nodes according to $p_t^o$. Thus we get the sub set of $\mathcal{O}_t$:  $\mathcal{O}'_t=\{\mathbf{o}_{t,m}\}^M_{m=1}$ and the importance scores of the selected objects ${p^o_t}'\in\mathbb{R}^{1 \times M}$. 
\emph{Secondly}, we obtain the edge set $\mathcal{A}_t'$ via keeping the corresponding edges between selected objects:
\begin{equation}
    \mathcal{A}_t'=\{e_{t,ij} | \mathbf{o}_{t,i} \in \mathcal{ O}_t' \land \mathbf{o}_{t,j} \in \mathcal{O}_t' \land e_{t,ij} \in \mathcal{A}_t \}.
    \label{eq:induced edges}
\end{equation}
Therefore, the node feature and the adjacency matrix of $M$-order graph $\mathcal{G}_t'$ are $O'_t\in\mathbb{R}^{M \times N_o}$ and $A_t'\in\mathbb{R}^{M \times M}$.

Since navigation is a long-term planning task, RLM should also consider future navigation decisions. 
Thus we customise a reinforcement learning strategy to guide the subgraph sampling process. The reward for the object's node selection is based on the navigation action feedback from environments as shown in Figure~\ref{fig:pipline}(b). At each step $t$, the agent's action reward is defined as: (\romannumeral1) If the agent's distance from destination decreases, the reward $r_t$ will be $1$. Otherwise, it will be $-1$. (\romannumeral2) When the agent successfully finishes the navigation task, it will gain a large reward $r_T=4$, and if the agent stops at the wrong position, it will get a negative reward value $r_T=-2$. 
The navigation reward can be seen as the reward for the RLM. 
At each step $t$, the critic network $critic(\cdot)$ estimates the state value $v_t$, and each action reward $r_t$ for RLM is:
\begin{equation}
    r_t = \gamma^{(T-t)}v_T + r_t,\quad v_t = critic(h_t^o),
\end{equation}
where $\gamma$ is the decay rate, $T$ denotes the total steps of the trajectory. Therefore, the advantage of each action reward against state value is $\varsigma_t = r_t-v_t$. The action of subgraph miner is to choose $M$ objects from $\mathcal{O}_t$, and the objective functions based on A2C~\cite{mnih2016asynchronous} is formulated as:
\begin{equation}
   \left\{\begin{matrix}
   \begin{aligned}
    \small
    \mathcal{L}_{sa} &= \sum^{T}_{t=0}\sum^{M}_{m=0}-\varsigma_t\log {p^o_t}'_{,m}; \\
    \mathcal{L}_{sc} &= \sum^{T}_{t=0}\varsigma_t^2; \\
    \mathcal{L}_{sd} &=\sum^{T}_{t=0}\sum^{K}_{k=0}-p^o_{t,k}\log p^o_{t,k};\\
    \mathcal{L}_s   & = \lambda_1\mathcal{L}_{sa} + \lambda_2\mathcal{L}_{sc} + \lambda_3\mathcal{L}_{sd},
   \end{aligned}
   \end{matrix}\right.
\end{equation}
where $\mathcal{L}_{sa}$ optimises the object selection, $\mathcal{L}_{sc}$ optimises the critic model, and $\mathcal{L}_{sd}$ aims to avoid the object importance scores $p^o_t$ degeneration to uniform distribution. $\lambda_i(i=1,2,3)$ is the weighting factor that controls the relative importance of each term.
\begin{figure}
   \begin{center}
      \includegraphics[width=0.98\linewidth]{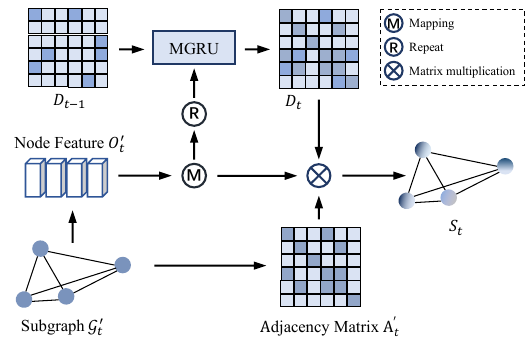}
   \end{center}
      \caption{ Illustration of our SEM module. SEM takes the layout graph $\mathcal{G}'_t$ from RLM to update the layout memory $D_t$ and generate the structured navigation state $S_t$.}\vspace{-5mm}
   \label{fig:SEM}
\end{figure}
\subsection{Structured Evolving Module}
\label{seg:SEM}

The object-level layout information is contained in the generated layout graph. However, the navigation is a sequential process, where the layout graph dynamically changes over time. Thue we propose the Structured Evolving Module (SEM) to dynamically handle the layout graph $\mathcal{G}'_t$, to update the navigation state $S_t$ at each step $t$. As shown in Figure \ref{fig:SEM}, we adopt a learnable matrix $D_t$ to record the structured layout memory. We leverage the matrix version of GRU (MGRU)~\cite{pareja2020evolvegcn} to handle $D_t \in \mathbb{R}^{N \times N}$, where $N=100$ is a hyperparameter. 
The $D_t$ is updated by the node feature $O'_t$ of the subgraph from RLM. We first repeat and mapping $O'_t \in  \mathbb{R}^{M\times N_o}$ to produce a square matrix $Q_t\in\mathbb{R}^{N \times N}$ ($M<N$). Then the updating process of ${D_{t-1}}\rightarrow {D_{t}}$ can be formulated as:
\begin{equation}
   \left\{\begin{matrix}
   \begin{aligned}
Z_t &= \sigma(W_z Q_t+U_zD_{t-1}+B_z); \\
R_t &= \sigma(W_r Q_t+U_r  D_{t-1} + B_r); \\
\tilde{D}_t &= \tanh(W_h  Q_t + U_h (R_t D_{t-1})+B_h); \\
D_t&=(1-Z_t) \circ D_{t-1} + Z_t \circ \tilde{D}_t,
   \end{aligned}
   \end{matrix}\right.
\end{equation}
where $\circ$ denotes the Hadamard product, $\sigma$ represents the sigmoid function and the trainable parameters are all square matrix of order $N$. 
For the initialisation of $D_t$, since there is no history layout information at the first step, the $D_0$ should condition on the language instruction:
\begin{equation}
 D_0=\sigma( I_0 w_L^\top W_3 + B_i), 
\end{equation}
where $I_0 \in \mathbb{R}^{N \times 1}$, $B_i \in \mathbb{R}^ {N\times N}$,$W_3 \in \mathbb{R}^{N_h \times N}$ are trainable parameters. $w_L \in \mathbb{R}^{N_h \times 1}$ is the sentence level feature.

The obtained history structure memory $D_t$ guides a graph convolution on the current layout graph to produce the current structured navigation state: 
\begin{equation}
S_t = \tanh(A_t' \sigma(O'_{t}W_m )D_t),    
\end{equation}
where $W_m \in \mathbb{R}^{N_o \times N}$ is learnable. Due to the interaction between the subgraph $\mathcal{G}'_t$ from RLM and the layout memory matrix $D_t$, the $S_t$ carries the layout relation of the current step with historical memory. 
We use the language context-aware feature $\tilde{h}_t$ to conduct attention pooling on the $S_t$ to build a global representation $h^s_t$ for action selection:
\begin{equation}
    h_t^s = \sigma( W_4[\tilde{s}_t,h_t]),\quad \tilde{s}_t = attn(\tilde{h}_t,S_t),
\end{equation}
where $W_4 \in \mathbb{R}^{(N_h+N_o)\times N_h}$ is learnable. 

Finally, $h_t^s$ serves as the action selector in the action prediction module. For simplicity, the action prediction module computes the attention similarity between the $h_t^s$ and the feature $ f_t = [f_{t,i}]_{i=0}^{K_t}$ that represents the candidate views:
\begin{equation}
    p_t = softmax(h^s_t W_5 f_t),
\end{equation}
where $W_5 \in \mathbb{R}^{N_h \times N_v}$ is the trainable parameter, $N_v$ is the dimension of visual feature, and $f_{t,0}$ is a zero vector representing the stop action. $p_t$ is the action probability distribution at step $t$. The agent samples/selects a candidate view based on $p_t$ and move to the viewpoint corresponding to that view.
\subsection{Training Loss} 
Our training objective is composed of three different parts: the imitation learning loss $\mathcal{L}_I$, the reinforcement learning for action prediction $\mathcal{L}_r$ and the subgraph extraction loss $\mathcal{L}_s$ mentioned before.
During the training process, we implement the student-force training strategy, in which the agent takes action that is predicted by itself. With the $a^g_t$ as the shortest path to the destination, the imitation learning loss is defined as:
\begin{equation}
\small
    \mathcal{L}_I = \lambda\sum^T_{t=0}a^g_t \log P_t.
\end{equation}
The reinforcement learning uses the A2C~\cite{mnih2016asynchronous} strategy:
 \begin{equation}
 \small
 \mathcal{L}_r = \sum^T_{t=0}-a_t\log(P_t)adv_t,
\end{equation}
where the $adv_t$ is the advantage at step $t$ in A2C algorithm. 
The overall training loss can be defined as:
\begin{equation} 
\small
    \mathcal{L} = \mathcal{L}_s + \lambda_I \mathcal{L}_I + \lambda_r\mathcal{L}_r,
\end{equation}
where $\lambda_I$, $\lambda_r$ are coefficients for balancing the loss terms.
\section{Experiments}
\begin{table*}[htb]
   \begin{center}
      \resizebox{\linewidth}{!}{
\definecolor{Gray}{gray}{0.94}
\begin{tabular}{lrrrrrr>{\columncolor{Gray}}r>{\columncolor{Gray}}rrr>{\columncolor{Gray}}r>{\columncolor{Gray}}r}
			\toprule 
			\multicolumn{1}{c}{} & \multicolumn{4}{c}{Val Seen} & \multicolumn{4}{c}{Val Unseen} & \multicolumn{4}{c}{Test Unseen} \\
			\cmidrule(r){2-5} \cmidrule(r){6-9} \cmidrule(r){10-13}
			\multicolumn{1}{c}{Agent} & \multicolumn{1}{c}{TL} & \multicolumn{1}{c}{NE$\downarrow$} & \multicolumn{1}{c}{SR$\uparrow$} & \multicolumn{1}{c}{SPL$\uparrow$} & \multicolumn{1}{c}{TL} & \multicolumn{1}{c}{NE$\downarrow$} & \multicolumn{1}{c}{SR$\uparrow$} & \multicolumn{1}{c}{SPL$\uparrow$} & \multicolumn{1}{c}{TL} & \multicolumn{1}{c}{NE$\downarrow$} & \multicolumn{1}{c}{SR$\uparrow$} & \multicolumn{1}{c}{SPL$\uparrow$} \\
			\midrule
			Random                & 9.58  & 9.45 & 0.16 & -    & 9.77  & 9.23 & 0.16 & -    & 9.89  & 9.79 & 0.13 & 0.12 \\
			Human                 & -     & -    & -    & -    & -     & -    & -    & -    & 11.85 & 1.61 & 0.86 & 0.76 \\
			\midrule
			\emph{Pretrain Model:}\\
			PRESS~\cite{li2019press}
			& 10.57 & 4.39 & 0.58 & 0.55 & 10.36 & 5.28 & 0.49 & 0.45 & 10.77 & 5.49 & 0.49 & 0.45 \\
			PREVALENT~\cite{hao2020prevelant}
			& 10.32 & 3.67 & 0.69 & 0.65 & 10.19 & 4.71 & 0.58 & 0.53 & 10.51 & 5.30 & 0.54 & 0.51 \\
			VLNBERT (init. OSCAR)~\cite{hong2020recurrent}	& 10.79 & 3.11 & 0.71 & 0.67 & 11.86 & 4.29 & 0.59 & 0.53 & 12.34 & 4.59 & 0.57 & 0.53 \\
			VLNBERT (init. PREVALENT)~\cite{hong2020recurrent} &  11.13 & 2.90 & 0.72 & 0.68 & 12.01 & 3.93 & 0.63 & 0.57 & 12.35 & 4.09 & 0.63 & 0.57 \\
			VLNBERT+REM~\cite{liu2021vision}
			& 10.88 & 2.48 & 75.4 & 71.8 & 12.44 &3.89 & 0.64 & 0.60 &13.11 & 3.87 & 0.65 & 0.59 \\ 
			ORIST (init. UNITER)~\cite{qi2021road}
			& - & - & - & - & 10.90 & 4.72 & 0.57 & 0.51 & 11.31 & 5.10 & 0.57 & 0.52 \\
			\midrule
			\emph{w/ Data Augmentation:}\\
			Speaker-Follower~\cite{fried2018speaker}
			& -     & 3.36 & 0.66 & -    & -     & 6.62 & 0.35 & -    & 14.82 & 6.62 & 0.35 & 0.28 \\
			SM~\cite{ma2019self}
			& -     & {3.22} & 0.67 & 0.58 & -     & 5.52 & 0.45 & 0.32 & 18.04 & 5.67 & 0.48 & 0.35 \\
			RCM+SIL~\cite{wang2019rcm}
			& 10.65 & 3.53 & 0.67 & -    & 11.46 & 6.09 & 0.43 & -    & 11.97 & 6.12 & 0.43 & 0.38 \\
			Regretful~\cite{ma2019regretful}
			& -     & 3.23 & 0.69 & 0.63 & -     & 5.32 & 0.50 & 0.41 & 13.69 & 5.69 & 0.48 & 0.40 \\
			IL+RL+REM~\cite{liu2021vision}
			& 10.18 & 4.61 & 0.58 & 0.55 & \textbf{9.40} & 5.59 & 0.48 & 0.44 & \textbf{9.81} & 5.67 & 0.48 & 0.45 \\
			SSM~\cite{wang2021structured} 
			& 14.70 & \textbf{3.10} & \textbf{0.71} & 0.62  & 20.70 & 4.32 & \textbf{0.62} & 0.45 & 20.40 & 4.57 & 0.61 & 0.46 \\
			VLNBERT (no init.)~\cite{hong2020recurrent} & 9.78 & 3.92 & 0.62 & 0.59 & 10.31 & 5.10 & 0.50 & 0.46 & 11.15 & 5.45 & 0.51 & 0.47 \\
			EnvDrop~\cite{tan2019learning}
			& 11.00 & 3.99 & 0.62 & 0.59 & 10.70 & 5.22 & 0.52 & 0.48 & 11.66 & 5.23 & 0.51 & 0.47 \\
			EnvDrop+REM~\cite{liu2021vision}
			& 11.13 & 3.14 & 0.70 & 0.66 & 14.84 & 4.99 & 0.53 & 0.48 & 10.73 & 5.40 & 0.54 & 0.50 \\  
			AuxRN~\cite{zhu2020vision}
			& -     & 3.33 & {0.70} & \textbf{0.67} & -     & 5.28 & 0.55 & 0.50 & -     & 5.15 & 0.55 & 0.51 \\
			RelGraph~\cite{hong2020language}                  & \textbf{10.13} & 3.47 & 0.67 & 0.65 & 9.99  & 4.73 & 0.57 & 0.53 & 10.29 & 4.75 & 0.55 & 0.52 \\
			NvEM~\cite{an2021neighborview}  & 11.09 & 3.44 & 0.69 & 0.65 & 11.83 & {4.27} & {0.60} & {0.55} & 12.98 & {4.37} & {0.58} & {0.54} \\
		    \textbf{EnvDrop+SEvol} & 12.55     &  3.70   &  0.61   &  0.57    &  14.67   &  4.39   &  0.59 &  0.53   &  14.30    &  3.70  &   0.59  &  0.55   \\  
		    \textbf{NvEM+SEvol} 
		    & 11.97     &  3.56   &  0.67   &  0.63    &  12.26   &  \textbf{3.99}   &  \textbf{0.62} &  \textbf{0.57}   &  13.40    &  \textbf{4.13}  &   \textbf{0.62}  &  \textbf{0.57}   \\   
            \midrule
            \emph{w/o Data Augmentation:}\\
			Student-Forcing~\cite{anderson2018vision} \
			& 11.33 & 6.01 & 0.39 & -     & 8.39 & 7.81 & 0.22 & -    & \textbf{8.13} & 7.85 & 0.20 & 0.18 \\
			RPA~\cite{wang2018look} 
			& \textbf{8.46} & 5.56 & 0.43 & -     & \textbf{7.22} & 7.65 & 0.25 & -    & 9.15 & 7.53 & 0.25 & 0.23 \\
			Regretful~\cite{ma2019regretful}
			& -    & 3.69 & 0.65 & \textbf{0.59}  &  -   & 5.36 & 0.48 & 0.37 & -    &  -   & -    & - \\
			EGP~\cite{deng2020evolving}
			& -    & -    & -    & -     &  -   & 5.34 & 0.52 & 0.41 & -    &  -   & -    & - \\
			EnvDrop~\cite{tan2019learning}
			& 10.10 & 4.71 & 0.55 & 0.53  & 9.37 & 5.49 & 0.47 & 0.43 & -    &  -   & -    & - \\ 
			Active Perception~\cite{wang2020active} 
			& 19.80 & \textbf{3.35} & \textbf{0.66} &  0.51 & 19.90 & 4.40 & 0.55 & 0.40 & 21.0 & 4.77 & 0.56 &0.37 \\
			SSM~\cite{wang2021structured} 
			& 13.50 & 3.77 & 0.65 & 0.57 & 18.90 & 4.88 & 0.56 & 0.42 & 18.50 & 4.66 & 0.57 & 0.44 \\ 
		    {\textbf{EnvDrop+SEvol}}
			& 12.33 & 4.19 & 0.59 & 0.54 & 13.59 & 4.72 & 0.55 & 0.49 & 15.48   & 4.52   & 0.57    & 0.52 \\		    
		    {\textbf{NvEM+SEvol}}
			& 12.07 & 3.96 & 0.63 & \textbf{0.59} & 12.45 & \textbf{4.15} & \textbf{0.61} & \textbf{0.55} & 13.10    & \textbf{4.25}    & \textbf{0.60}    & \textbf{0.55} \\
			\bottomrule
	\end{tabular}
      }
   \end{center}\vspace{-2mm}
   \caption{\textbf{Comparisons on R2R dataset}. Comparison of single run performance with the state-of-the-art methods on R2R. The proposed SEvol boosts the performance in terms of all the key metrics on the unseen environments.
   }
   \label{tab:sota}
\end{table*}

\begin{table*}[t]
	\centering
	\resizebox{0.9\textwidth}{!}{
 
{\begin{tabular}{lrrrrrrrrrrrr}
			\toprule 
			\multicolumn{1}{c}{} & \multicolumn{6}{c}{Val Seen} & \multicolumn{6}{c}{Val Unseen} \\
			\cmidrule(r){2-7} \cmidrule(r){8-13}
			\multicolumn{1}{c}{Agent} & \multicolumn{1}{c}{NE$\downarrow$} & \multicolumn{1}{c}{SR$\uparrow$} & \multicolumn{1}{c}{SPL$\uparrow$} & \multicolumn{1}{c}{CLS$\uparrow$} & \multicolumn{1}{c}{nDTW$\uparrow$} & \multicolumn{1}{c}{sDTW$\uparrow$} & \multicolumn{1}{c}{NE$\downarrow$} & \multicolumn{1}{c}{SR$\uparrow$} & \multicolumn{1}{c}{SPL$\uparrow$} & \multicolumn{1}{c}{CLS$\uparrow$} & \multicolumn{1}{c}{nDTW$\uparrow$} & \multicolumn{1}{c}{sDTW$\uparrow$}\\
			\midrule
			EnvDrop~\cite{tan2019learning} & - & 0.52 & 0.41 & 0.53 & - & 0.27 & - & 0.29 & 0.18 & 0.34 & - & 0.09 \\
			RCM-b~\cite{grabriel2019general}    
			& -    & -    & -    & -    & -    &  -   & -    & 0.29 & 0.21 & 0.35 & 0.30 & 0.13 \\
			OAAM~\cite{Qi2020object} & - & \textbf{0.56} & 0.49 & \textbf{0.54} & - & 0.32 & - & 0.31 & 0.23 & 0.40 & - & 0.11 \\
			RelGraph~\cite{hong2020language}  & \textbf{5.14} & 0.55 & \textbf{0.50} & 0.51 & 0.48 & \textbf{0.35 }& 7.55 & 0.35 & 0.25 & 0.37 & 0.32 & 0.18 \\
			NvEM~\cite{an2021neighborview}  & 5.38 & 0.54 & 0.47 & 0.51 & \textbf{0.48} & \textbf{0.35} & \textbf{6.85} & 0.38 & 0.28 & \textbf{0.41} & \textbf{0.36} & \textbf{0.20} \\
			\midrule
			\textbf{NvEM+SEvol} &5.77  & 0.50  & 0.40 & 0.48 & 0.45 & 0.30 & 6.90 & \textbf{0.39} & \textbf{0.29} &\textbf{0.41}& \textbf{0.36}&\textbf{0.20} \\
			\bottomrule
	\end{tabular}}
}
\caption{\textbf{Comparisons on R4R dataset}. Comparison of single run performance with the state-of-the-art methods on R4R. }
	\label{tab:r4r}
\end{table*}

\begin{table*}[t]
   \begin{center}
      \resizebox{0.85\linewidth}{!}{
         
\begin{tabular}{cc|c|c|c|cccc||ccccc}
  \hline
  & \multirow{2}{*}{Name} &   \multirow{2}{*}{SEM} &\multirow{2}{*}{RLM}  & \multirow{2}{*}{Aug$_{bp}$} & \multicolumn{4}{c||}{Val-Seen} & \multicolumn{4}{c}{Val-Unseen}  \\
  &                       &                              &                              &                             & SR$\uparrow$         & NE$\downarrow$           & TL$\downarrow$   & SPL$\uparrow$   & SR$\uparrow$   & NE$\downarrow$     & TL$\downarrow$   & SPL$\uparrow$   \\ \shline
  &  NvEM~\cite{an2021neighborview}*       &                             &                              &                              & 0.61    &   4.25   &  \textbf{{10.85}}   & 0.58  & 0.57   &  4.62  & \textbf{{11.24}} & 0.51 \\     
    &   \#1                  & \checkmark                  &                    &
  & 0.62    &  {3.79}   &  12.16   & 0.58  & 0.60   &  4.31  & 12.41  & 0.54 \\
  &   \#2                  &   \checkmark                & \checkmark                   &
  & {0.63}    &   3.96   &  12.07   & {0.59}  & {0.61}   &  {4.15}  & 12.45  & {0.55} \\  \hline
  &   \#3                  & \checkmark                    &                 & \checkmark 
  & 0.64    &   3.79   &  11.66   & 0.59  & 0.61   &  4.14  & 12.40  & 0.56 \\
  &   \#4                  & \checkmark                  & \checkmark                   &\checkmark
		    &\textbf{0.67}    & \textbf{3.56}   &  11.97     &  \textbf{0.63}    &  \textbf{0.62}  &  \textbf{3.99}   &   12.26   &  \textbf{0.57} \\  
  \shline
\end{tabular}

      }
   \end{center}\vspace{-1mm}
   \caption{\textbf{Ablations.} The performance is gradually improved with the continuous addition of proposed modules, especially on val-unseen. The reproduction result of NvEM~\cite{an2021neighborview} \textit{w/o data augmentation} is shown at the first line. Experiments confirm the effectiveness of RLM/SEM modules under both the \textit{w/} and \textit{w/o data augmentation} setup. Note that Aug$_{bp}$ denotes data augmentation.}
   \label{tab:ablations1}
\end{table*}

\begin{table*}
   \begin{center}
    {
   
\subfloat[\textbf{Capacity of Layout Memory.} The Dimension of $D_t$ in SEM. Model with $dim(D_t)=100$ performs best on val-unseen. \label{tab:ablation:reasoning}]{
   \tablestyle{2.5pt}{1.05}\begin{tabular}{c|x{24}x{24}x{24}|x{24}x{24}x{24}}
      \multirow{2}{*}{Dim($D_t$)} & \multicolumn{3}{c|}{Val-Seen} & \multicolumn{3}{c}{Val-Unseen}             \\ \cline{2-7}
  & SR$\uparrow$   & NE$\downarrow$  & SPL$\uparrow$ & SR$\uparrow$ & NE$\downarrow$ & SPL$\uparrow$ \\\shline
    \scriptsize \emph{--} & 0.61       & 4.25       & 0.58     & 0.57   & 4.62      & 0.51     \\\hline
    \scriptsize \emph{50} & \textbf{0.65 }      & \textbf{3.72 }      & \textbf{0.60 }    & 0.58   &  4.43     & 0.52\\ 
    \scriptsize \emph{100} & 0.63       & 3.96       & 0.59     & \textbf{0.61}   & \textbf{4.15}      & \textbf{0.55 }     \\
    \scriptsize \emph{200}  & 0.63       & 3.89       & 0.58     & 0.60   & 4.33      & 0.54          \\      
\end{tabular}}\hspace{3mm}
\subfloat[\textbf{Subgraph Size.} The size of subgraph selected by RLM. The model performs best on val-unseen when the subgraph with size of 5. \label{tab:ablation:knowledge}]{
   \tablestyle{2.5pt}{1.05}\begin{tabular}{c|x{24}x{24}x{24}x{24}|x{24}x{24}x{24}x{24}}
      \multirow{2}{*}{\scriptsize \emph{Num}}  & \multicolumn{4}{c|}{Val-Seen}                     & \multicolumn{4}{c}{Val-Unseen}               \\ \cline{2-9}
                                                   & SR$\uparrow$  &  NE$\downarrow$   & TL$\downarrow$  & SPL$\uparrow$     & SR$\uparrow$ & NE$\downarrow$   & TL$\downarrow$ & SPL$\uparrow$ \\ \shline
      \scriptsize \emph{0}                         & 0.61        & 4.25        &     \textbf{10.85}      & 0.58          & 0.57      & 4.62        & \textbf{11.24}       & 0.51      \\ \hline
      \scriptsize \emph{3}                         & 0.60      & 4.21        & 11.26       & 0.56         & 0.57   & 4.45       & 11.67   & 0.52       \\
      \scriptsize \emph{5}                        & \textbf{0.63}      & \textbf{3.96}    & 12.07     & \textbf{0.59}         & \textbf{0.61}     & \textbf{4.15 }          & 12.45       & \textbf{0.55}        \\
      \scriptsize \emph{20}                        & 0.59  & 4.03       & 12.28      &  0.55      & 0.59     & 4.33       & 12.64     & 0.54      \\ 
   \end{tabular}}\hspace{3mm}

   }
   \end{center}
   \vspace{-2mm}
   \caption{\textbf{Ablations.} We mainly focus on the key metrics for each ablation setting. Note that the results in this table are obtained under the \emph{w/o data augmentation} setting.}
   \label{tab:ablations2}\vspace{-1mm}
\end{table*}

\begin{table}
   \begin{center}\resizebox{\linewidth}{!}{

\begin{tabular}{c|x{24}x{24}x{24}|x{24}x{24}x{24}}
      \multirow{2}{*}{$m(x)$} & \multicolumn{3}{c|}{Val-Seen} & \multicolumn{3}{c}{Val-Unseen}             \\ \cline{2-7}
  & SR$\uparrow$   & NE$\downarrow$  & SPL$\uparrow$ & SR$\uparrow$ & NE$\downarrow$ & SPL$\uparrow$ \\\shline
    $x$ & \textbf{0.63 }      &  3.96     & \textbf{0.59}  & \textbf{0.61}   & 4.15   & \textbf{0.55} \\  \hline
    $e^x$ & 0.59       &  4.26     & 0.55  & 0.57   &   4.53    & 0.53     \\ \hline
    $1$  & 0.61       &    \textbf{3.88}   &  0.57   & 0.58  &    \textbf{3.34}   &     0.53      \\      
\end{tabular}
   }
   \end{center}\vspace{-2mm}
   \caption{Performances with different edge weight generation functions $d(m(x))$, where $x$ is heading difference between two objects.}
   \label{tab:edge}
\end{table}

\subsection{Experimental Setup}
\label{sec:exp}
\vspace{1mm}
\noindent\textbf{Datasets.} We adopt Room-to-Room(R2R) dataset~\cite{anderson2018vision} and Room-for-Room dataset~\cite{sotp2019acl} to validate our method. In the R2R dataset, there are $7,189$ trajectories and three natural language instructions describe each. It is divided into train, validation seen (val-seen), validation unseen (val-unseen), and test sets. The environments of val-unseen and test are not in the train set.  The R4R dataset contains longer and twistier paths, which is the extended version of R2R. The R4R dataset is divided into train,  validation seen and validation unseen sets. The R2R and R4R are built on the Matterport3D simulator~\cite{anderson2018vision,chang2017matterport3d}, which consists of $10,567$ panoramic views in 90 real word indoor environments.

\noindent\textbf{Evaluation Metrics.}
To compare with the existing methods, we report the commonly used evaluation metrics on the R2R dataset, \ie, Trajectory Length (TL), Navigation Error (NE), Success Rate (SR), and Success Rate weighted by Path Length (SPL). 
Following~\cite{sotp2019acl,ilharco2019general,an2021neighborview}  three more evaluations are used for R4R, \ie, normalized Dynamic Time Warping (nDTW), Success rate weighted normalized Dynamic Time Warping (sDTW) and Coverage weighted by Length Score (CLS).

\noindent\textbf{Implementation Details.} We use a 2-layer bi-directional LSTM with a cell size of 256 in each direction for the language encoder and use GloVe~\cite{pennington2014glove} for word embedding. We use the CLIP-ViT-B-32 as our visual encoder.
The LSTM decoder is with a cell size of 512.
The SEvol can be seen as an additional branch for the LSTM based VLN model. More specifically, $h^s_t$ produced by SEvol can replace $h_t$, the output of the LSTM.
At each step, we detect 400 objects to build $\mathcal{G}_t$. The object detector is Faster R-CNN~\cite{ren2015faster} trained on Visual Genome Dataset~\cite{krishna2017visual} which classifies the 100 most frequent objects appearing in the instructions and environments.
The REM module selects 5 objects from $\mathcal{G}_t$ to induces the subgraph $\mathcal{G}'_t$. The edge weight function over the objects' angle difference $x$ is $d(x) = \frac{\pi}{\pi+6x}$.
Following~\cite{tan2019learning}, we apply the two-stage training strategy on R2R dataset. At the first stage, we train the agent on the train set. At the second stage, the agent is trained with the back translation based data augmentation method~\cite{fried2018speaker} 
The learning rate is $10^{-4}$. The weighting factor's values are $\lambda_1=0.2$, $\lambda_2=0.1$, $\lambda_3=0.01$, $\lambda_I=0.2$ and $\lambda_r=1$. The batch size is 64. We train our model for $8,000$ iterations for stage $1$, and $20,000$ iterations for stage $2$ and select the model with the best performance on val-unseen set.

\subsection{Comparison with State-of-the-art Methods}
\noindent\textbf{R2R Dataset}. 
As shown in Table~\ref{tab:sota}, under the \textit{w/ data argumentation} setting, our best model reach the same SR with the previous best one~(SSM~\cite{wang2021structured}), but we achieve a significant improvement in SPL (from $0.45$ to $0.57$) on val-unseen. The NvEM+SEvol surpasses NvEM~\cite{an2021neighborview} with $+2\%$ in both SR and SPL. Besides, we achieve $+4\%$ and $+3\%$ absolute improvement in SR and SPL than NvEM~\cite{an2021neighborview} on test set. SSM~\cite{wang2021structured} sacrifices the navigation efficiency to achieve $0.61$ SR on test set, and we achieve higher SR ($+1\%$) and higher efficiency ($+11\%$ in SPL). The SEvol also boosts the performance of EnvDrop for a large margin. Under the \textit{w/o data argumentation} setting, as shown in Table~\ref{tab:sota}, we achieve the best performance in terms of SR and SPL on both val-unseen and test sets, and it is higher than the previous best methods in a large margin. Even compared with the previous state-of-the-art NvEM~\cite{an2021neighborview} trained with augmented data, the NvEM+SEvol performs better on both val-unseen and test set in terms of SR and SPL. 

\vspace{0.5mm}
\noindent\textbf{R4R Dataset}. 
As shown in Table~\ref{tab:r4r}, we compare the proposed method with the current state-of-the-art methods on the R4R benchmark. The NvEM+SEvol achieves the best performance on most of the metrics on val-unseen set. And it suppresses the NvEM model on both SR and SPL by $1\%$ and achieves the same accuracy on the trajectory fidelity metrics. The experiments show that the proposed SEvol also works on samples with longer navigation trajectories.

\subsection{Ablation Study}
In this section, we evaluate the effectiveness of the key components of our model, \ie, RLM and SEM. Note that the base model in our experiments is NvEM~\cite{an2021neighborview}. We reproduce the experiment without data argumentation and add the key components of our model step by step.

\vspace{0.5mm}
\noindent\textbf{Structured Evolving Module (SEM)}. As shown in Table~\ref{tab:ablations1}, compare with base agent, `\#1' with SEM lifts SR and SPL from ($0.57$, $0.51$) to ($0.60$, $0.54$) on val-unseen. It confirms that the layout memory and structured state are conducive to navigation. Moreover, after training with augmented data (`\#3'), the model still benefits from structured state maintaining.

\vspace{0.5mm}
\noindent\textbf{Reinforced Layout clues Miner (RLM).} As shown in Table~\ref{tab:ablations1}, comparing to `\#1', `\#2' confirms the effectiveness of RLM by boosting SR and SPL from $0.60$, $0.54$ to $0.61$, $0.65$. The advantage remains after training with augmentation data, as shown in `\#4'. It proves that selecting suitable objects will help the layout memory to keep useful information of the environment and produce better structured navigation state.

\vspace{0.5mm}
\noindent\textbf{Capacity of Structure Memory.} As shown in Table~\ref{tab:ablations2}(a), we investigate how the capacity of the layout memory in SEM impacts the performance by changing the dimension of $D_t$. Note that $dim=0$ represents the base model without the SEvol. We can observe that no matter the dimension of $D_t$, the model with SEM performs better than the baseline model. When the dimension rises from $50$ to $100$, SR on val-unseen increases to $0.61$ from $0.58$, which illustrates that the larger capacity of the structure memory has better expressiveness. The model performs best when $dim=100$. SR on val-unseen decreases to $0.60$ when $dim=200$, which indicates that high-dimensional $D_t$ increases the convergence difficulty.

\vspace{0.5mm}
\noindent\textbf{Subgraph Size.} As shown in Table~\ref{tab:ablations2}(b), we investigate how the node number of $\mathcal{G'}_t$ selected by RLM influences the performance. We compare the models with different sizes of subgraph, \ie, node number of $3$, $5$ and $20$. The model with the subgraph $\mathcal{G'}_t$ size of $5$ achieves the best performance. When the subgraph size increases to $20$, SR on val-unseen decreases, which indicates that a large layout graph brings more noise.

\vspace{0.5mm}
\noindent\textbf{Layout Graph Construction.} We investigate the different spatial layout graph construction methods. Specifically, we choose mapping functions $m(\cdot)$ to influence the decay speed of edge weights over $x$, the difference of angle between objects. As shown in Table~\ref{tab:edge}, $x$ represents the angle difference between two objects. When $m(x)=x$, the model has best performance, which is better than $m(x)=1$. When $m(x)=e^x$, the model has the lowest SR 
on val-unseen. Thus the edge weight should not decrease too fast since the relationships among objects in the different directions is useful for navigation.

\subsection{Visualisation}

\begin{figure}
\vspace{2mm}
   \begin{center}
      \includegraphics[width=1\linewidth]{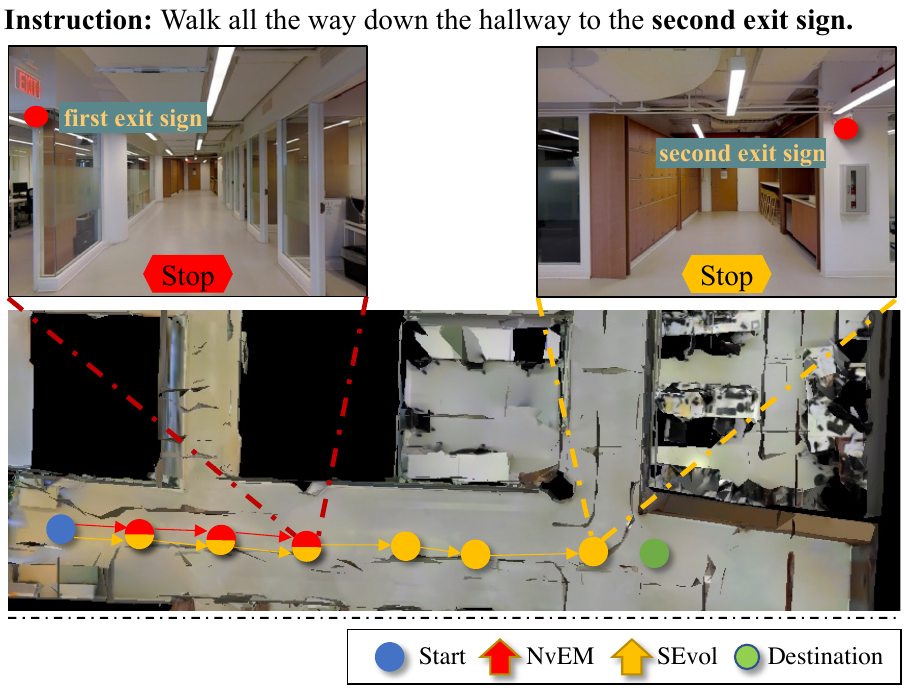}
   \end{center}
      \caption{The decision divergence between the SEvol and baseline model NvEM~\cite{an2021neighborview}. The proposed SEvol obtains more layout clues than the baseline model.}
   \label{fig:vis_2}
\end{figure}

To demonstrate that the history layout information maintained by the SEvol model assists the agent's decision-making process, we visualise the trajectories generated by the agent with or without the SEvol model. 
In Figure~\ref{fig:vis_2}, the agent with SEvol does not stop at the first exit sign, then gets to the right destination. However, the agent without SEvol stops at the wrong position. Due to the lack of the object layout history information, it can not distinguish the first and the second sign. 

\section{Conclusion and Discussion}

The object-level layout information is critical for the Vision-and-Language Navigation task. To utilise the layout information, we propose a novel model for vision-and-language navigation, named the Structured state-Evolution model (SEvol), which maintains a structured memory and evolves a structured navigation state. We offer RLM to extract the navigation-related subgraph from the environment and SEM to maintain a layout memory for evolving structured navigation. 
Extensive experiments demonstrate the effectiveness of our proposed methods. 
The proposed SEvol substantially improves VLN performance on the R2R dataset, proving the importance of layout information in the VLN task. 
We believe that this work will bring new insights to the research around vision-and-language navigation.

\noindent\textbf{Limitations.} Although SEvol outperforms previous state-of-the-art methods by a large margin, the layout information in SEvol only contains orientation relations between objects. Thus how to represent the layout relation containing more valuable clues is worth studying in the future. Besides, due to the insufficient environment data on the R2R dataset, models trained on R2R may have bias over some typical environments. Thus constructing more diverse datasets is a primary task for the VLN community.
In the future, we will continue the research on the VLN task around the limitations.

{\small
\bibliographystyle{ieee_fullname}
\bibliography{egbib}
}

\end{document}